\documentclass[10pt,twocolumn,letterpaper]{article}
\usepackage{multirow}
\usepackage{btas}
\usepackage{times}
\usepackage{epsfig}
\usepackage{graphicx}
\usepackage{amsmath}
\usepackage{amssymb}
\usepackage{caption}
\usepackage{enumitem}

\btasfinalcopy 

\ifbtasfinal\fi
\begin{document}

\title{Generative Adversarial Network-based Synthesis of Visible Faces from Polarimetric Thermal Faces}

\author{
He  Zhang$^1$, Vishal M. Patel$^1$, Benjamin S. Riggan$^2$, and Shuowen Hu$^2$\\
\\
$^1$ Rutgers, The State University of New Jersey, Piscataway, NJ 08854\\
$^2$ U.S. Army Research Laboratory, 2800 Powder Mill Rd., Adelphi, MD 20783
\\
{\tt\small {\{he.zhang92,vishal.m.patel\}}@rutgers.edu, \{benjamin.s.riggan.civ, shuowen.hu\}@mail.mil}
}

\maketitle
\thispagestyle{empty}

\begin{abstract}
The large domain discrepancy between faces captured in polarimetric (or conventional) thermal and visible domain makes cross-domain face recognition quite a challenging problem for both  human-examiners and computer vision algorithms. Previous approaches utilize a two-step procedure (visible feature estimation and  visible image reconstruction) to synthesize the visible image given the corresponding polarimetric thermal image. However, these  are regarded as two disjoint steps and hence may hinder the performance of visible face reconstruction. We argue that joint optimization would be a better way to reconstruct  more photo-realistic images for both computer vision algorithms and human-examiners to examine. To this end, this paper proposes a Generative Adversarial Network-based Visible Face Synthesis (GAN-VFS) method to synthesize more photo-realistic visible face images from their corresponding polarimetric images.  To ensure that the encoded visible-features contain more semantically meaningful information in reconstructing the visible face image, a guidance sub-network is involved into the training procedure. To achieve photo realistic  property while preserving discriminative characteristics for the reconstructed outputs, an identity loss combined with the perceptual loss are optimized in the framework. Multiple experiments evaluated on different experimental protocols demonstrate that the proposed method achieves state-of-the-art performance.  
\end{abstract}


\section{Introduction}
Face recognition is one of the most widely studied topics in computer vision.  In recent years,
methods based on deep convolutional neural networks (CNNs) have shown impressive performance improvements for face recognition problems \cite{vggface,face_net,face_residue}.   Even though these methods are able to address many challenges such as the low-resolution, pose variation and illumination variation to some extent, they are specifically designed for recognizing face images that are collected near-visible spectrum.  They often do not perform well on the face images captured from other domains such as polarimetric \cite{ ijcb_datasets2, btas_2016, ijcb_datasets}, infrared \cite{Klare_NIR, SWIR_TIFS} or millimeter wave \cite{mmW_ISBA2017} due to significant phenomenological differences as well as a lack of sufficient training data. 
\begin{figure}[t]
	\centering
		\begin{minipage}{.15\textwidth}
			\centering
			\includegraphics[width=1\textwidth]{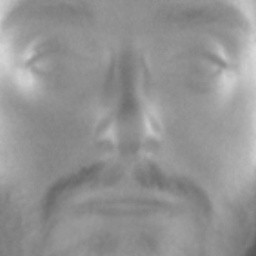}
			\captionsetup{labelformat=empty}
			\captionsetup{justification=centering}
			\caption*{(a)}
	\end{minipage}
	\begin{minipage}{.15\textwidth}
		\centering
		\includegraphics[width=1\textwidth]{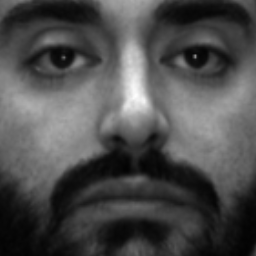}
		\captionsetup{labelformat=empty}
		\captionsetup{justification=centering}
		\caption*{(b)}
	\end{minipage}
	\begin{minipage}{.15\textwidth}
		\centering
		\includegraphics[width=1\textwidth]{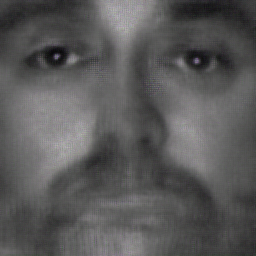}
		\captionsetup{labelformat=empty}
		\captionsetup{justification=centering}
    	\caption*{(c)}
	\end{minipage}\\
	\begin{minipage}{.15\textwidth}
			\centering
			\includegraphics[width=1\textwidth]{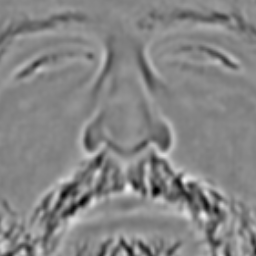}
			\captionsetup{labelformat=empty}
			\captionsetup{justification=centering}
			\caption*{(d)}
	\end{minipage}
	\begin{minipage}{.15\textwidth}
		\centering
		\includegraphics[width=1\textwidth]{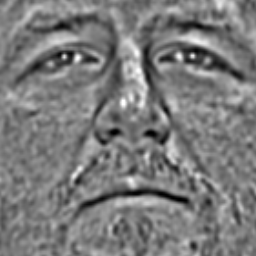}
		\captionsetup{labelformat=empty}
		\captionsetup{justification=centering}
		\caption*{(e)}
	\end{minipage}
	\begin{minipage}{.15\textwidth}
		\centering
		\includegraphics[width=1\textwidth]{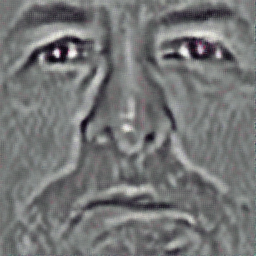}
		\captionsetup{labelformat=empty}
		\captionsetup{justification=centering}
    	\caption*{(f)}
	\end{minipage}	
	\vskip -8pt
	\caption{Sample results of our proposed method, GAN-VSF. (a) Input Polarimetric image.  (b) Ground truth visible image.  (c)  Estimated visible image using GAN-VSF from the image shown in (a).  (d), (e), and (f) are the DoG filtered images corresponding to the images shown in (a), (b), and (c), respectively.}  \label{fig:over}
\end{figure}
Distributional change between thermal and visible images makes thermal-to-visible face recognition very challenging.  Various methods have been developed in the literature for bridging this gap, seeking to develop a cross-domain face recognition algorithm \cite{cross_domain3,face_cross1,face_cross2,cross_5}.  In particular, recent works have proposed using the polarization-state information of thermal emissions to enhance the performance of cross-spectrum face recognition \cite{ ijcb_datasets2, btas_2016, ijcb_datasets}.  It has been shown that polarimetric-thermal images capture geometric and textural details of faces that are not present in the conventional thermal facial imagery \cite{ijcb_datasets}.    As a result, the use of polarization-state information can perform better than using only conventional thermal imaging for face recognition.

Previous approaches have attempted to improve the cross-domain face recognition performance, however, it is still of great importance to guarantee that the human examiners can identify whether the given polarimetric and the visible image share the same identity or not.  Consider the polarimetric face image shown in Figure~\ref{fig:over}(a).  The corresponding visible image is shown in Figure~\ref{fig:over}(b).  As can be seen from these images, it is extremely difficult for either human-examiners or existing face  recognition systems to determine whether these two images share the same identity.   Hence, methods that can automatically generate high quality visible images from their corresponding polarimetric images are needed.

One such method was recently proposed in \cite{btas_2016}.    However, it suffers from the following drawbacks.  1) The procedures of visible feature extraction and image reconstruction are not jointly optimized and hence may degrade the reconstruction performance.  2) The recovered images are not photo realistic and hence may hinder the verification performance.   In this work,  we take a major step towards  addressing these two issues.  In order to optimize the overall algorithm jointly, a generative adversarial network (GAN) based approach with a guidance sub-network is proposed to generate more sharp visible faces.  Furthermore, to generate more photo realistic images and to make sure that the generated images contain highly discriminative information, an identity loss combined with the perceptual loss is included in the optimization.  Quantitative and qualitative experiments evaluated on  four different polarimetric-to-visible settings demonstrate that the proposed method can achieve state-of-the-art performance as compared to previous methods.   

Samples results of the proposed Generative Adversarial Network-based Visible Face Synthesis (GAN-VFS) method are shown in Figure~\ref{fig:over}.  Given the polarimetric face images shown in Figure~\ref{fig:over}(a), our method automatically synthesizes a visible image shown in Figure~\ref{fig:over}(c), which is very close to the ground truth visible image shown in Figure~\ref{fig:over}(b).  Images in Figure~\ref{fig:over}(d)-(f) show the difference-of-Gaussian (DoG) filtered images corresponding to the images shown in Figure~\ref{fig:over}(a)-(c).

\begin{figure*}[t]
\centering
\includegraphics[width=0.9\textwidth]{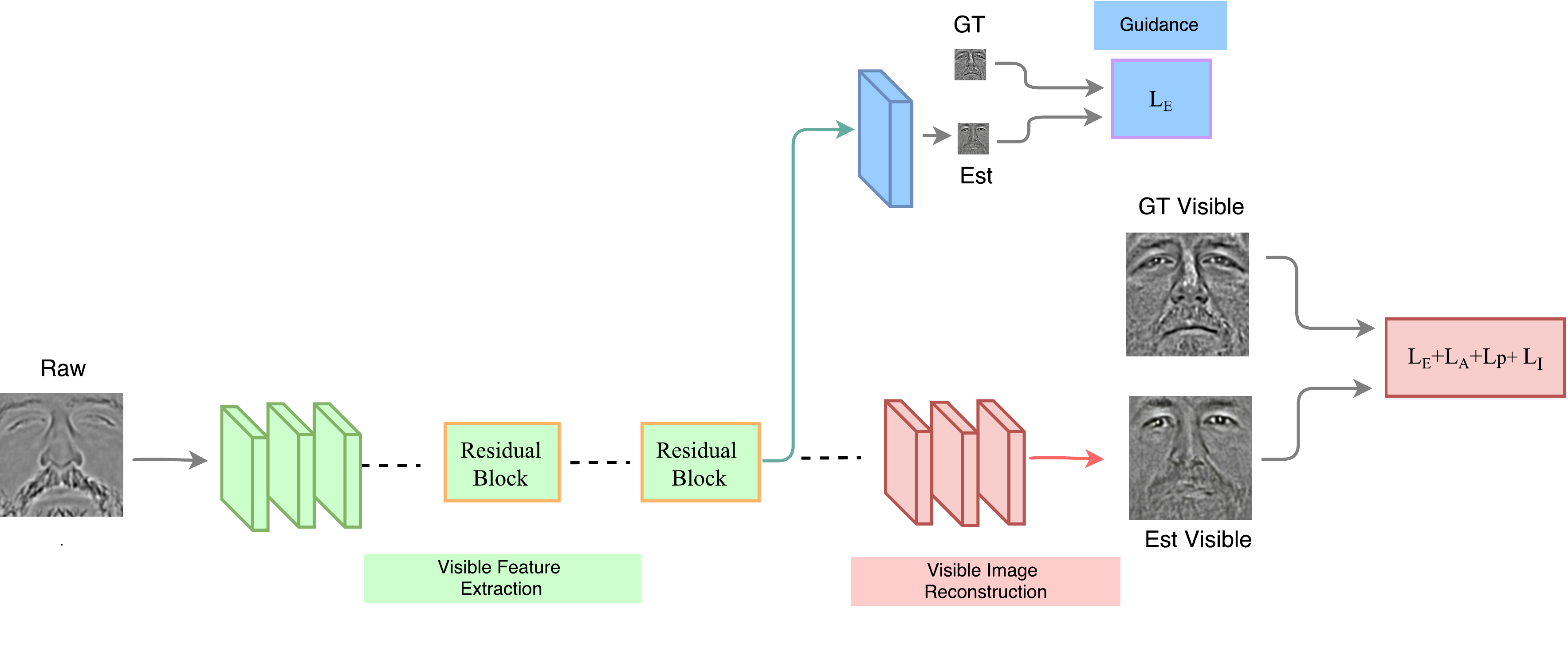}
\vskip-20pt \caption{Overview of the proposed GAN-VFS method.  It contains three modules. (a) Visible feature extraction module, (b) Guidance sub-network and (c) Visible image reconstruction module. Firstly, the visible feature is extracted from the raw polarimetric image.  Then, to make sure that the learned feature can better reconstruct the visible image, a guidance sub-network is involved into the optimization. Finally, the guided feature is used to reconstruct the photo realistic visible image using the combination of different losses.}
\label{fig:overview}
\end{figure*}

\section{Generative Adversarial Networks (GANs)}
Generative Adversarial Networks were proposed by Goodfellow et al. in \cite{GAN} to synthesize realistic images by effectively learning the distribution of training images.  The authors adopted a game theoretic min-max optimization framework to simultaneously train two models: a generative model, $G$, and a discriminative model, $D$.   The success of GANs in synthesizing realistic images has led to researchers exploring the adversarial loss for numerous low-level vision  applications such as style transfer \cite{li2016precomputed}, image in-painting \cite{pathak2016context}, image to image translation \cite{reed2016generative}, image super-resolution \cite{GAN_sisr}, image de-raining \cite{derain_2017_zhang} and crowd counting \cite{croudgan}. Inspired by the success of these methods, we propose to use the adversarial loss to learn the distribution of visible face images for their accurate estimation.

\section{Proposed Method}
Instead of optimizing the two procedures (visible feature estimation and visible image reconstruction) separately, a new unified GAN-based framework is proposed in this section. In the following sub-sections, we discuss these important parts in detail starting with the GAN objective function followed by details of the proposed network and the loss functions.

\subsection{GAN Objective Function}
In order to learn a good generator $G$ so as to fool the learned discriminator $D$ and  to make the discriminator $D$ good enough to distinguish synthesized visible image from real ground truth,  the proposed method alternatively updates $G$ and $D$ following the structure proposed in \cite{GAN_pix2pix2016,derain_2017_zhang}. Given an input polarimetric image $X$, conditional GAN aims to learn a mapping function to generate output image $Y$ by solving the following optimization problem:
\begin{equation}\label{eq:GAN1}
\begin{split}
\min_G \max_D \quad & \mathbb E _{X\sim p_{data(X)} }[\log (1- D(X, G(X)))]+\\
&\mathbb E _{X\sim p_{data(X,Y)}}[\log D(X,Y)].
\end{split}
 \end{equation}

\subsection{Network Overview}
The process of transforming  the image from polarimetric domain to visible domain can be regarded as a pixel-level image regression problem.  Basically, the transforming procedure can be divided into two separated processes as discussed in \cite{btas_2016}. Firstly,  a set of features that can aid the process of reconstructing visible image are extracted from the given polarimetric images.  Then, an optimization procedure is proposed to reconstruct the corresponding visible face images via those learned visible features.  Even though the previous method \cite{btas_2016} achieves very good performance in transforming the image from polarimetric domain to visible domain via the proposed two steps, the proposed two steps are not jointly learned and optimized.  In other words, the proposed two-step method has to rely on the assumption that the learned features contain semantically meaningful information in reconstructing the visible face images. 

To overcome this issue and to relax that assumption, a unified GAN-based synthesis network that can directly learn an end-to-end mapping between the polarimetric image and its corresponding visible image is proposed.  The whole network contains an encoder-decoder structure, where the learned visible features can be regarded as the outputs of the encoder part and input for the decoder part. To guarantee the reconstructability  of the  encoded features and to make sure that the leaned features contain semantic information, a guidance sub-network \cite{deep_supervision}, is introduced at the end of the visible feature extraction part.  The overall network architecture is shown in Figure~\ref{fig:overview}. 

To overcome the side effect of blurry results brought by the traditional Euclidean loss ($L_E$ loss) \footnote{We use $L_E$ loss to represent the Euclidean loss throughout the paper. } and to discriminate the generated visible face images from their corresponding ground truth, a GAN structure is deployed. Even though GAN's structure can generate more reasonable results compared to the tradition $L_E$ loss, as will be shown later, the results generated by the traditional GAN contain undesirable facial artifacts, resulting in a less photo realistic image. To address this issue and meanwhile generate visually pleasing results, the  perceptual loss is included in our work, where the perceptual loss is evaluated on pre-trained VGG-16 models, as discussed in \cite{perceptual_loss,hang_style}. 

As the ultimate goal of the our proposed synthesis method is to guarantee that human examiners can identify the person given his synthesized face images, it is also important to involve the discriminative information into consideration.  Similar to the perceptual loss, we propose an identity loss that is evaluated on a certain layer of the fine-tuned VGG-Polar model. The VGG-Polar model is fine-tuned using the visible images with its corresponding labels from the Polarimetric-Visible dataset \cite{ijcb_datasets2}.

\subsection{Network Structure}
Figure~\ref{fig:overview} gives an overview of the proposed synthesis framework.  We adopt an encoder-decoder structure as the basis in the generator part.  Basically, the proposed generator can be divided into two parts. Firstly, a set of convolutional layers with stride 2 combined with a set of residual blocks \cite{perceptual_loss} are regarded as visible feature estimation part. Specifically, the residual blocks are composed of two convolutional layers with 3$\times3$ kernels and 64 feature maps followed by batch-normalization layers \cite{batch_norm} and PReLU \cite{prelu} as the activation function. Then, a set of transpose convolutional layers with stride 2 are denoted as the visible image reconstruction procedure.   To make sure that the transformed features contain enough semantic information, a guided sub-part is enforced in the network. Meanwhile, to make the generated visible face images indistinguishable from the ground truth visible face images, a CNN-based differentiable discriminator is used as a guidance to guide the generator in generating better visual results. For the discriminator, we use PatchGANs \cite{GAN_pix2pix2016} to discriminate whether the given images are real or fake.

The structure of the proposed discriminator network is as follows: 

\noindent \emph{CB(32)-CBP(64)-CBP(128)-CBP(256)-CBP(256)-C(1)-Sigmoid},\\

\noindent where $C$ represents the covolutional layer,  $P$ indicates Prelu \cite{prelu} and $B$ denotes as batch-normalization \cite{batch_norm}. The number in the bracket represents the number of output feature maps of the corresponding layer.
\begin{figure*}[!]
	\centering
	\begin{minipage}{.135\textwidth}
			\centering
			\includegraphics[width=1\textwidth]{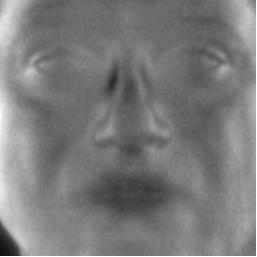}
			\captionsetup{labelformat=empty}
			\captionsetup{justification=centering}
	\end{minipage}
	\begin{minipage}{.135\textwidth}
		\centering
		\includegraphics[width=1\textwidth]{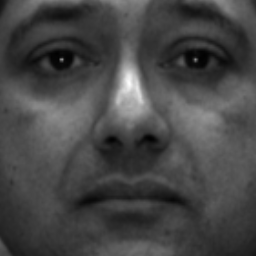}
		\captionsetup{labelformat=empty}
		\captionsetup{justification=centering}
	\end{minipage}
	\begin{minipage}{.135\textwidth}
		\centering
		\includegraphics[width=1\textwidth]{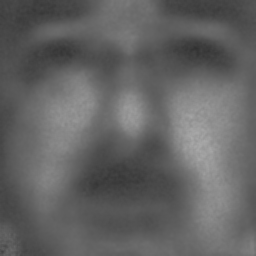}
		\captionsetup{labelformat=empty}
		\captionsetup{justification=centering}
	\end{minipage}	
	\begin{minipage}{.135\textwidth}
			\centering
			\includegraphics[width=1\textwidth]{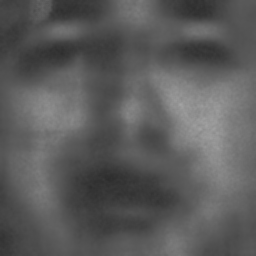}
			\captionsetup{labelformat=empty}
			\captionsetup{justification=centering}
	\end{minipage}
	\begin{minipage}{.135\textwidth}
		\centering
		\includegraphics[width=1\textwidth]{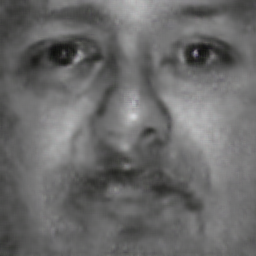}
		\captionsetup{labelformat=empty}
		\captionsetup{justification=centering}
	\end{minipage}
	\begin{minipage}{.135\textwidth}
		\centering
		\includegraphics[width=1\textwidth]{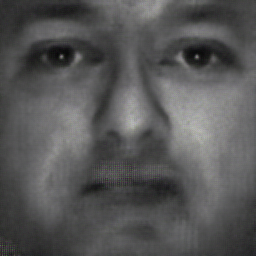}
		\captionsetup{labelformat=empty}
		\captionsetup{justification=centering}
	\end{minipage}
	\begin{minipage}{.135\textwidth}
		\centering
		\includegraphics[width=1\textwidth]{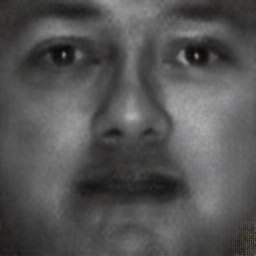}
		\captionsetup{labelformat=empty}
		\captionsetup{justification=centering}
	\end{minipage}\\  
	\begin{minipage}{.135\textwidth}
			\centering
			\includegraphics[width=1\textwidth]{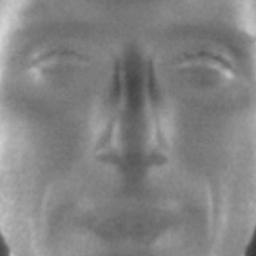}
			\captionsetup{labelformat=empty}
			\captionsetup{justification=centering}
			\caption*{(a)}
	\end{minipage}
	\begin{minipage}{.135\textwidth}
		\centering
		\includegraphics[width=1\textwidth]{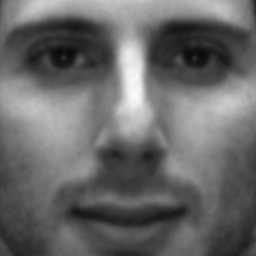}
		\captionsetup{labelformat=empty}
		\captionsetup{justification=centering}
		\caption*{(b)}
	\end{minipage}
	\begin{minipage}{.135\textwidth}
		\centering
		\includegraphics[width=1\textwidth]{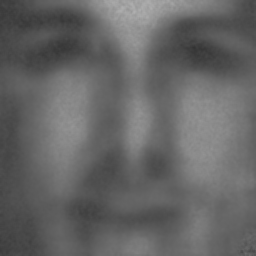}
		\captionsetup{labelformat=empty}
		\captionsetup{justification=centering}
    	\caption*{(c)}
	\end{minipage}	
	\begin{minipage}{.135\textwidth}
			\centering
			\includegraphics[width=1\textwidth]{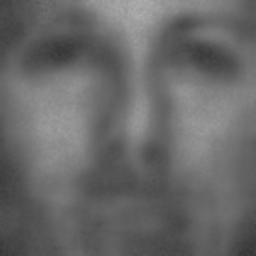}
			\captionsetup{labelformat=empty}
			\captionsetup{justification=centering}
			\caption*{(d)}
	\end{minipage}
	\begin{minipage}{.135\textwidth}
		\centering
		\includegraphics[width=1\textwidth]{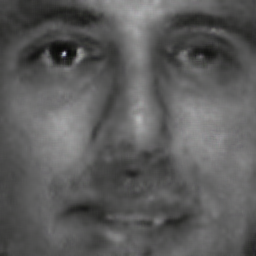}
		\captionsetup{labelformat=empty}
		\captionsetup{justification=centering}
		\caption*{(e)}
	\end{minipage}
	\begin{minipage}{.135\textwidth}
		\centering
		\includegraphics[width=1\textwidth]{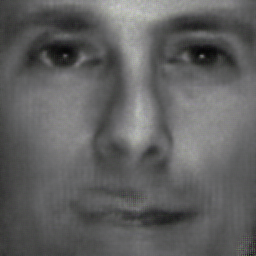}
		\captionsetup{labelformat=empty}
		\captionsetup{justification=centering}
    	\caption*{(f)}
	\end{minipage}
	\begin{minipage}{.135\textwidth}
		\centering
		\includegraphics[width=1\textwidth]{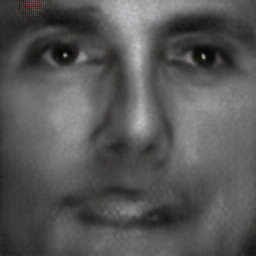}
		\captionsetup{labelformat=empty}
		\captionsetup{justification=centering}
    	\caption*{(h)}
	\end{minipage}
	\vskip -8pt\caption{Comparisons of visible face images synthesized with different experimental configurations. (a) Raw polarimetric images. (b) Ground truth visible images. (c) Reconstructed images with only $L_E$ loss.
	(d) Reconstructed images with $L_E$ loss and guidance $L_E$ loss. (e) Reconstructed images with GAN, $L_E$ and guidance $L_E$ loses.  (f) Reconstructed images with GAN, $L_E$, guidance $L_E$ and perceptual losses. (h) Reconstructed images using with all five losses. } \label{fig:ablation}
\end{figure*}

\subsection{Loss Functions} 
The proposed method contains the following loss functions: the Euclidean $L_{E(G)}$ loss  enforced on the recovered visible image, the $L_E$ loss enforced on the guidance part,  the adversarial loss to guarantee more sharp results, the perceptual loss to preserve more photo realistic details and the identity loss to preserve more distinguishable information for the outputs. The overall loss function is defined as follows

\begin{equation}
\label{eq:overall_loss}
L_{\text{all}} = L_{E} + L_{E(G)}+ \lambda_A L_{A} + \lambda_PL_{P} +\lambda_IL_{I} ,
\end{equation}
where $L_{E}$ denotes the Euclidean  loss, $L_{E(G)}$ denotes the Euclidean  loss on the guidance sub-network, $L_A$ represents the adversarial loss, $L_P$ indicates the perceptual loss and $L_I$ is the identity loss. Here, $\lambda_A$, $\lambda_P$ and $\lambda_I$ are the corresponding weights. 

The $L_E$ and the adversarial losses are defined as follows:
\begin{equation}
\label{eq:trans_loss_euc}
 L_{E}, L_{E(G)}= \frac{1}{WH}\sum_{w=1}^{W}\sum_{h=1}^{H} \|\phi_{G}({I})^{w,h}-Y_t^{w,h}\|_2,	\\
 \end{equation} 
 \begin{equation}
 \label{eq:trans_loss_adv}
   L_A= -\log(\phi_D(\phi_G({I})^{c,w,h}),
 \end{equation}
 where $I$ is the input polarimetric image, $Y_t$ is the ground truth visible image, $W\times H$ is the dimension of the input image, $\phi_{G}$ is the generator sub-network $G$ and $\phi_D$ is the discriminator sub-network $D$.

As the perceptual loss and the identity losses are evaluated on a certain layer of the given CNN model, both can be defined as follows:  
\begin{equation}
\label{eq:dehaze_loss_perc}
 L_{P,I}= \frac{1}{C_iW_iH_i}\sum_{c=1}^{C_i}\sum_{w=1}^{W_i}\sum_{h=1}^{H_i} \|V(\phi_E({I}))^{c,w,h}-V
 (Y_t)^{c,w,h}\|_2,
\end{equation}
 where $Y_t$ is the ground truth visible image, $\phi_{E}$ is the proposed generator, $V$ represents a non-linear CNN transformation and $C_i,W_i,H_i$ are the dimensions of a certain high level layer $V$, which differs for perceptual and identity losses.

\subsection{Implementation}
The entire network is trained on a Nvidia Titan-X  GPU using the Torch-7 framework \cite{torch}. We choose $\lambda_A=0.005$ for the adversarial loss,  $\lambda_P=0.8$  for the perceptual loss and  $\lambda_I=0.1$ for the identity loss. During training, we use ADAM \cite{adam_opt} as the optimization algorithm with learning rate  of $8\times 10^{-4}$ and batch size of  3 images.  All the pre-processed training samples are resized to $256\times 256$. The perceptual loss is evaluated on relu3-1 layers the in the pre-trained VGG \cite{vggface} model.  The identity loss is evaluated on the relu2-2 layer of the fine-tuned VGG-Polar model.

\section{Experimental Results}
In this section, we demonstrate the effectiveness of the proposed approach by conducting various experiments on a unique dataset that contains polarimetric and visible image pairs from 60 subjects  \cite{ijcb_datasets2}.  Basically, a polarimetric image, referred as  Stokes images is composed of three channels: S0, S1 and S2. Here, S0 represents the conventional thermal image, whereas S1 and S2 represent the horizontal/vertical and diagonal polarization-state information, respectively.  The dataset was collected at three distances: Range 1 (2.5 m), Range 2 (5 m), and Range 3 (7.5 m).   Following the protocol defined in  \cite{btas_2016}, we only use the data corresponding to range 1 (baseline and expression).   In particular, 30 subjects from range 1 are selected for training and the remaining 30 subjects are used for evaluation.  We repeat this process multiple times and report average results.  We evaluate the performance of different methods on the tightly cropped DoG filtered images \cite{btas_2016}.   We also conduct another set of experiments to directly transfer the polarimetric images to visible images. To summarize, our proposed network is evaluated on the following four protocols: 
\begin{enumerate}[nolistsep]
\item[(a)] Conventional thermal (S0) to Visible (Vis). \footnote{Conventional thermal (it is also referred as S0 in this paper.) measures the total intensity of thermal emissions.}
\item[(b)] Polarimetric thermal (Polar) to Visible (Vis).
\item[(c)] DoG of S0 (S0-DoG) to DoG of Visible (Vis-DoG).
\item[(d)] DoG of Polarimetric thermal (Polar-DoG) to DoG of Visible (Vis-DoG).
\end{enumerate}

\subsection{Ablation Study} 
In order to better demonstrate the improvements obtained by different modules in the proposed network, we perform an ablation study involving the following experiments: 
1) Polar to Visible estimation with only $L_E$ loss, 2) Polar to Visible estimation with $L_E$ loss and guidance $L_E$ loss, 3) Polar to Visible estimation with GAN, $L_E$ and guidance$L_E$ loss, (4) Polar to Visible estimation with GAN, $L_E$, guidance $L_E$ loss and perceptual loss, and finally (5) Polar to Visible estimation with all five losses. Due to the space limitations, the ablation study is only evaluated on the polar-to-visible image synthesis experiments. Two reconstruction results are shown in Figure~\ref{fig:ablation}. It can be observed from the results that the $L_E$ loss itself generates much blurry faces. Even though the results with $L_E$ guidance sub-network are slightly better with more details, they are still very blured and many high frequency details are missing. By involving the GAN structure in the proposed methods, more details are added to the results.  But it can be observed that GAN itself produces images with artifacts.    The introduction of the perceptual loss in the proposed framework is able to tackle the artifacts better and makes the results visually pleasing. Finally, the combination of all five losses can generate more reasonable results. 

\begin{figure}[htp!]
\centering
\includegraphics[width=0.4\textwidth]{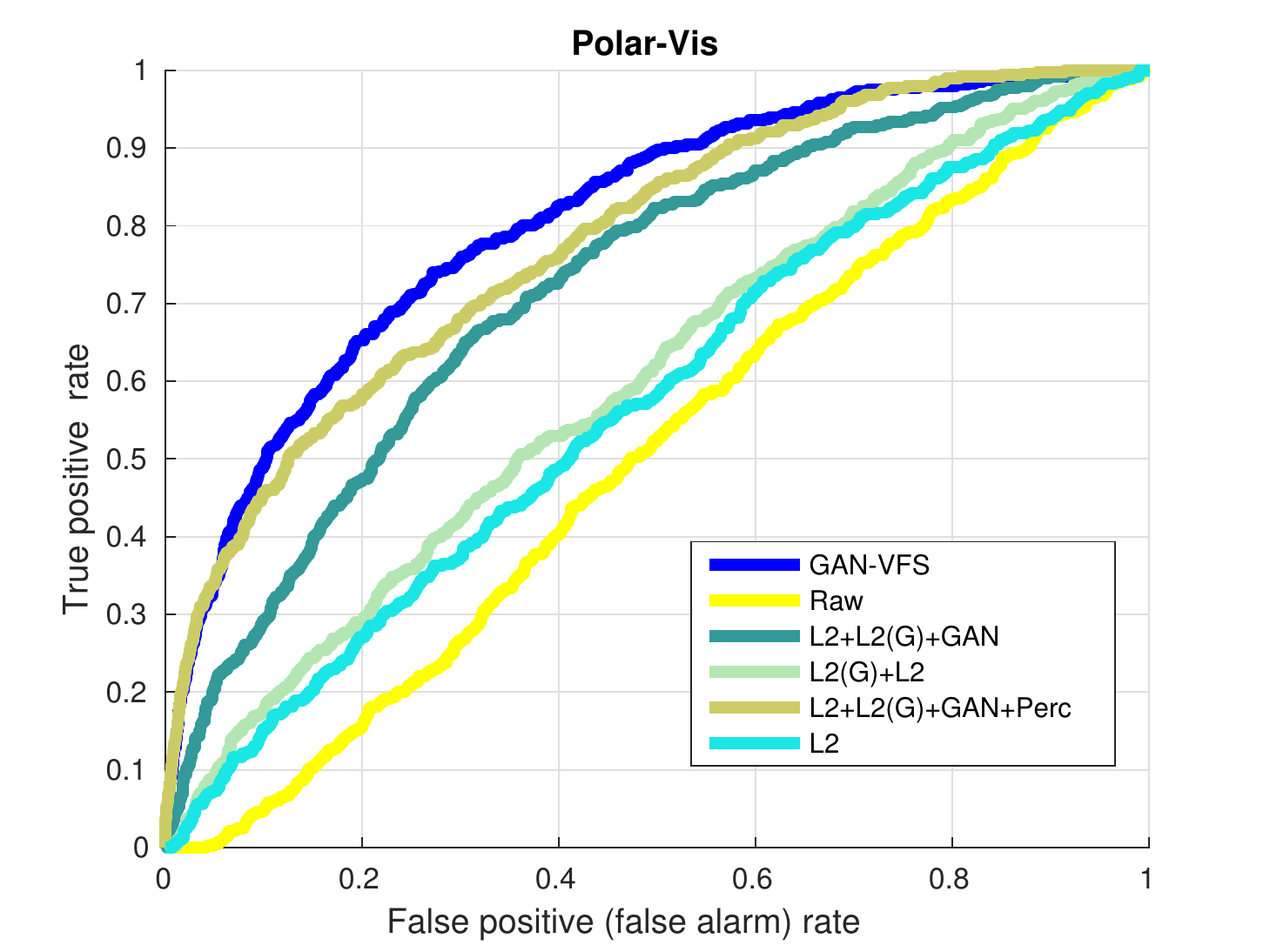}
 \vskip -6pt  \caption{The ROC curves corresponding to the ablation study. }
\label{fig:base}
\end{figure}

To better demonstrate the effectiveness of the different losses in the proposed methods, we plot the receiver operating characteristic (ROC) curves corresponding to the discussed five different network settings.   The results are shown in Figure~\ref{fig:base}. All the verification results are evaluated on the deep features extracted from the VGG-face model \cite{vggface} without fine-tuning.  From the ROC curves, it can be clearly observed that even though the identity loss does not produce visually different results, it can bring in more discriminative information.  

 \begin{figure*}[t]
\large \qquad S0-Vis (DoG)\qquad \quad Polar-Vis (DoG) \quad \quad\qquad S0-Vis \quad\qquad \qquad Polar-Vis\\
 	\centering
 	\begin{minipage}{.21\textwidth}
 			\centering
 			\includegraphics[width=.85\textwidth]{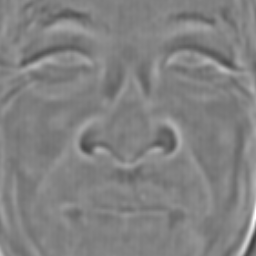}
 			\captionsetup{labelformat=empty}
 			\captionsetup{justification=centering}
 	\end{minipage}
 	\begin{minipage}{.21\textwidth}
 			\centering
 			\includegraphics[width=.85\textwidth]{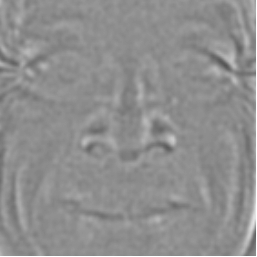}
 			\captionsetup{labelformat=empty}
 			\captionsetup{justification=centering}
 	\end{minipage}
 	\begin{minipage}{.21\textwidth}
 		\centering
 		\includegraphics[width=.85\textwidth]{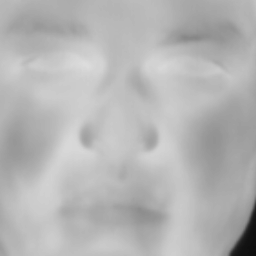}
 		\captionsetup{labelformat=empty}
 		\captionsetup{justification=centering}
 	\end{minipage}
 	\begin{minipage}{.21\textwidth}
 		\centering
 		\includegraphics[width=.85\textwidth]{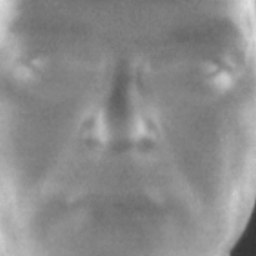}
 		\captionsetup{labelformat=empty}
 		\captionsetup{justification=centering}
 	\end{minipage}	\\\vskip+2pt
 	\large Raw\\\vskip+3pt
 	\begin{minipage}{.21\textwidth}
 			\centering
 			\includegraphics[width=.85\textwidth]{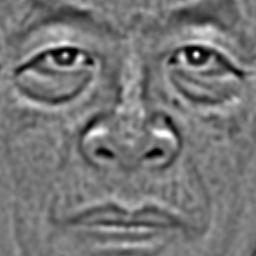}
 			\captionsetup{labelformat=empty}
 			\captionsetup{justification=centering}

 	\end{minipage}
 	\begin{minipage}{.21\textwidth}
 			\centering
 			\includegraphics[width=.85\textwidth]{figure//s0_46gt.png}
 			\captionsetup{labelformat=empty}
 			\captionsetup{justification=centering}

 	\end{minipage}
 	\begin{minipage}{.21\textwidth}
 		\centering
 		\includegraphics[width=.85\textwidth]{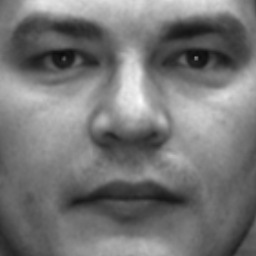}
 		\captionsetup{labelformat=empty}
 		\captionsetup{justification=centering}

 	\end{minipage}
 	\begin{minipage}{.21\textwidth}
 		\centering
 		\includegraphics[width=.85\textwidth]{figure//s0_25gt.png}
 		\captionsetup{labelformat=empty}
 		\captionsetup{justification=centering}

 	\end{minipage}\\\vskip+5pt
 	\large Ground Truth\\\vskip+3pt
  	\begin{minipage}{.21\textwidth}
  			\centering
  			\includegraphics[width=.85\textwidth]{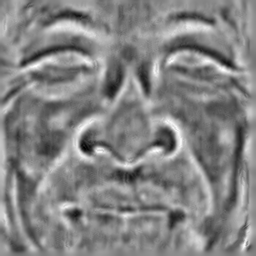}
  			\captionsetup{labelformat=empty}
  			\captionsetup{justification=centering}
  	\end{minipage}
  	\begin{minipage}{.21\textwidth}
  			\centering
  			\includegraphics[width=.85\textwidth]{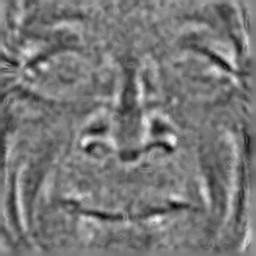}
  			\captionsetup{labelformat=empty}
  			\captionsetup{justification=centering}
  	\end{minipage}
  	\begin{minipage}{.21\textwidth}
  		\centering
  		\includegraphics[width=.85\textwidth]{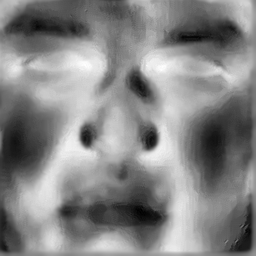}
  		\captionsetup{labelformat=empty}
  		\captionsetup{justification=centering}
  	\end{minipage}
  	\begin{minipage}{.21\textwidth}
  		\centering
  		\includegraphics[width=.85\textwidth]{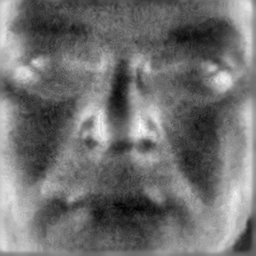}
  		\captionsetup{labelformat=empty}
  		\captionsetup{justification=centering}
  	\end{minipage}\\ 	\vskip+5pt
 	\large Mahendran \emph{et al.} \cite{inverting_cnn} \\\vskip+3pt
 	\begin{minipage}{.21\textwidth}
 			\centering
 			\includegraphics[width=.85\textwidth]{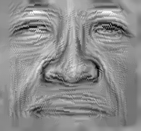}
 			\captionsetup{labelformat=empty}
 			\captionsetup{justification=centering}

 	\end{minipage}
 	\begin{minipage}{.21\textwidth}
 			\centering
 			\includegraphics[width=.85\textwidth]{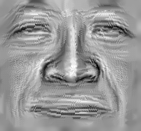}
 			\captionsetup{labelformat=empty}
 			\captionsetup{justification=centering}

 	\end{minipage}
 	\begin{minipage}{.21\textwidth}
 		\centering
 		\includegraphics[width=.85\textwidth]{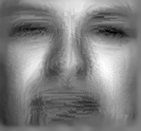}
 		\captionsetup{labelformat=empty}
 		\captionsetup{justification=centering}

 	\end{minipage}
 	\begin{minipage}{.21\textwidth}
 		\centering
 		\includegraphics[width=.85\textwidth]{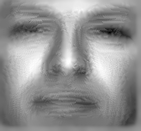}
 		\captionsetup{labelformat=empty}
 		\captionsetup{justification=centering}

 	\end{minipage}\\\vskip+5pt
 	\large Riggan \emph{et al.} \cite{btas_2016} \\\vskip+3pt
 	\begin{minipage}{.21\textwidth}
 			\centering
 			\includegraphics[width=.85\textwidth]{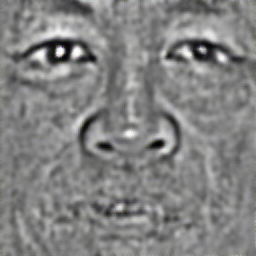}
 			\captionsetup{labelformat=empty}
 			\captionsetup{justification=centering}

 	\end{minipage}
 	\begin{minipage}{.21\textwidth}
 			\centering
 			\includegraphics[width=.85\textwidth]{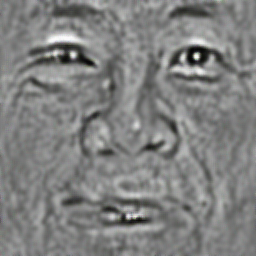}
 			\captionsetup{labelformat=empty}
 			\captionsetup{justification=centering}

 	\end{minipage}
 	\begin{minipage}{.21\textwidth}
 		\centering
 		\includegraphics[width=.85\textwidth]{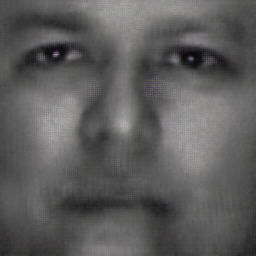}
 		\captionsetup{labelformat=empty}
 		\captionsetup{justification=centering}

 	\end{minipage}
 	\begin{minipage}{.21\textwidth}
 		\centering
 		\includegraphics[width=.85\textwidth]{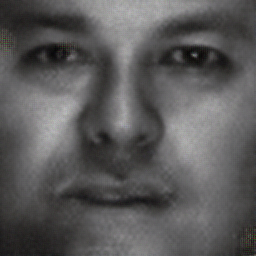}
 		\captionsetup{labelformat=empty}
 		\captionsetup{justification=centering}
 	\end{minipage}\\\vskip+2pt
 	\large GAN-VFS
 	\vskip -2pt\caption{Sample results corresponding to different methods on four protocols. } \label{fig:result}
 \end{figure*}

\subsection{Comparison with State-of-art Methods}  

To demonstrate the improvements achieved by the proposed method, it is compared against recent state-of-the-art methods on the discussed datasets.   In particular, we compare the performance of our method with that of the following two recent methods: Mahendran.  \emph{et al.} \cite{inverting_cnn}
and  Riggan. \emph{et al.} \cite{btas_2016}. 

As discussed above, four sets of experiments are conducted in evaluating the performance both qualitatively and quantitatively.   Peak signal-to-noise ratio (PSNR) and structural similarity (SSIM) \cite{ssim} are used to evaluate the synthesized image quality  performance of different methods.  

Results corresponding to the proposed method and recent state-of-the-art methods  on two sample images from the test dataset are shown in Figure~\ref{fig:result}. It can be observed from all four experiments (S0-Vis, Polar-Vis, S0-Vis (DoG) and Polar-Vis (DoG)) that the recovered results from our proposed method (last row) are more photo realistic and tend to recover more details. This can be clearly seen by comparing the eye regions of the recovered images.

The quantitative results evaluated using PSNR and SSIM are tabulated in Table~\ref{result_psnr} and Table~\ref{result_ssim}, respectively.  It
can be clearly observed that the proposed method is able to achieve superior quantitative performance compared the previous approaches.  These results highlight the significance of using a GAN-based approach to image synthesis.  

\begin{table}[htp!]
\centering
\caption{The average PSNR (dB) results corresponding to different methods.}
\label{result_psnr}
\resizebox{.48\textwidth}{!}{%
\begin{tabular}{|l|l|l|l|}
\hline
 & Mahendran. \emph{et al.} \cite{inverting_cnn} & Riggan. \emph{et al.} \cite{btas_2016} & \textbf{GAN-VFS} \\ \hline
S0-Vis & 9.58 & 12.09 & \textbf{17.11} \\ \hline
Poalr-Vis & 9.85 & 12.54 & \textbf{17.64} \\ \hline
S0-Vis (DoG) & 11.05 & 15.02 & \textbf{16.84} \\ \hline
Poalr-Vis (DoG) & 11.72 & 15.64 & \textbf{17.13} \\ \hline
\end{tabular}%
}
\end{table}

\begin{table}[htp!]
\centering
\caption{The average SSIM results corresponding to different methods.}
\label{result_ssim}
\resizebox{.48\textwidth}{!}{%
\begin{tabular}{|l|l|l|l|}
\hline
 & Mahendran. \emph{et al.} \cite{inverting_cnn} & Riggan. \emph{et al.} \cite{btas_2016} & \textbf{GAN-VFS} \\ \hline
S0-Vis & 0.2302 & 0.4342 & \textbf{0.5491} \\ \hline
Poalr-Vis & 0.2524 & 0.4423 & \textbf{0.5603} \\ \hline
S0-Vis (DoG) & 0.1785 & 0.2302 & \textbf{0.2604} \\ \hline
Poalr-Vis (DoG) & 0.1805 & 0.2404 & \textbf{0.2672} \\ \hline
\end{tabular}%
}
\end{table}

 \begin{figure*}[htp]
 	\centering
 	\begin{minipage}{.4\textwidth}
 			\centering
 			\includegraphics[width=1\textwidth]{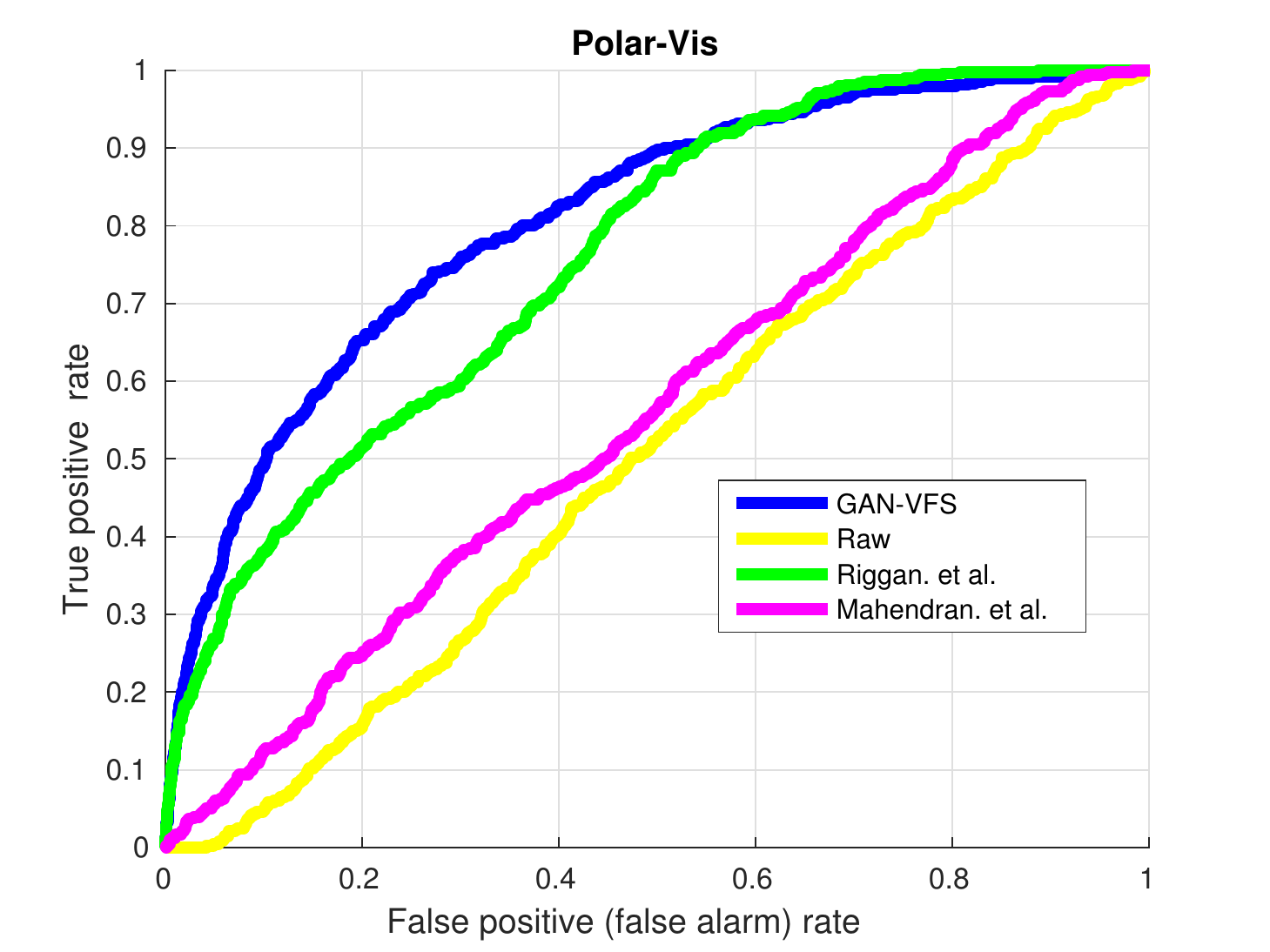}
 			\captionsetup{labelformat=empty}
 			\captionsetup{justification=centering}
 			\caption*{(a)}
 	\end{minipage}
 	\begin{minipage}{.4\textwidth}
 		\centering
 			\includegraphics[width=1\textwidth]{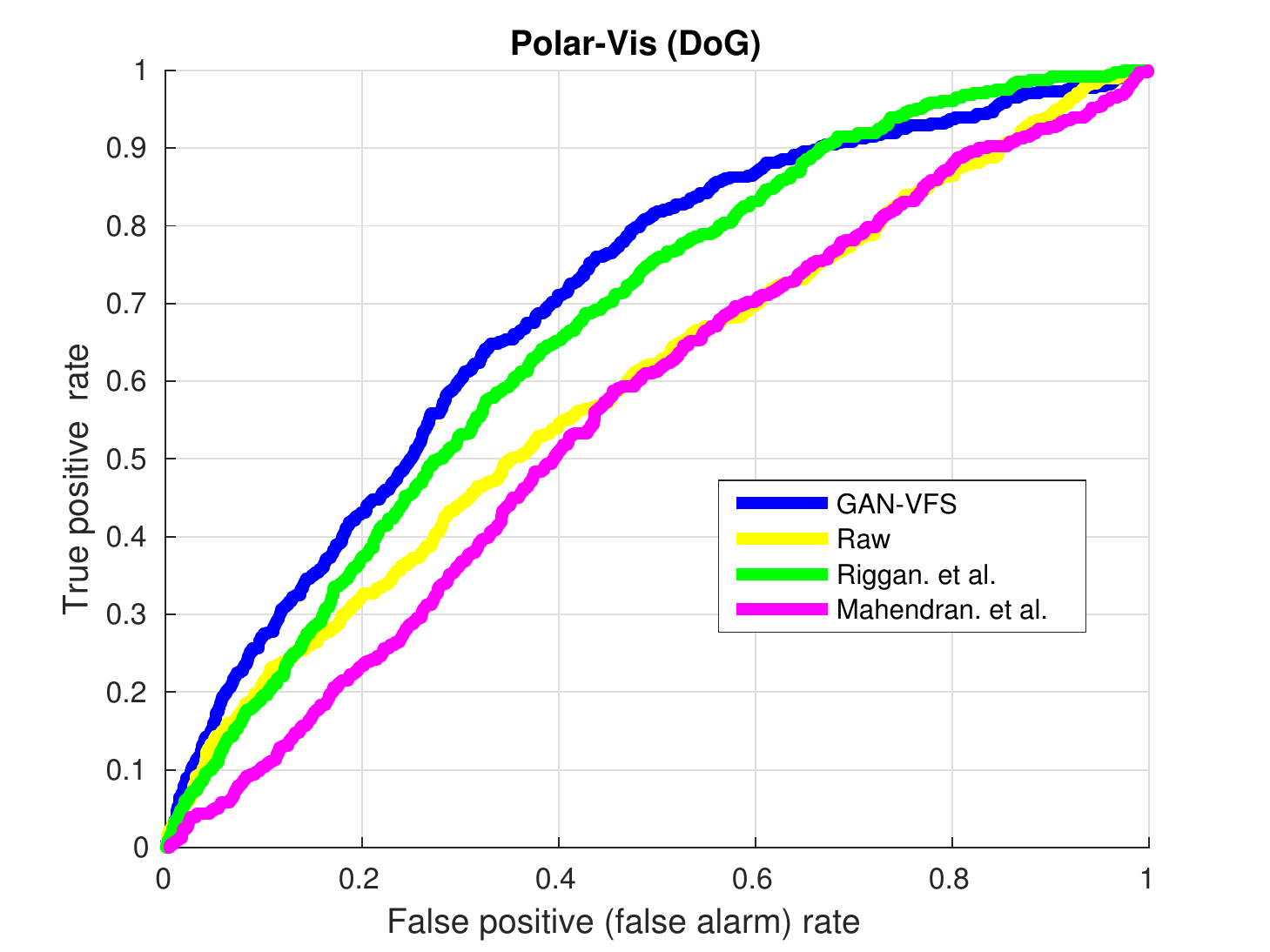}
 		\captionsetup{labelformat=empty}
 		\captionsetup{justification=centering}
 		\caption*{(b)}
 	\end{minipage}	\\
 	\begin{minipage}{.4\textwidth}
 			\centering
 			\includegraphics[width=1\textwidth]{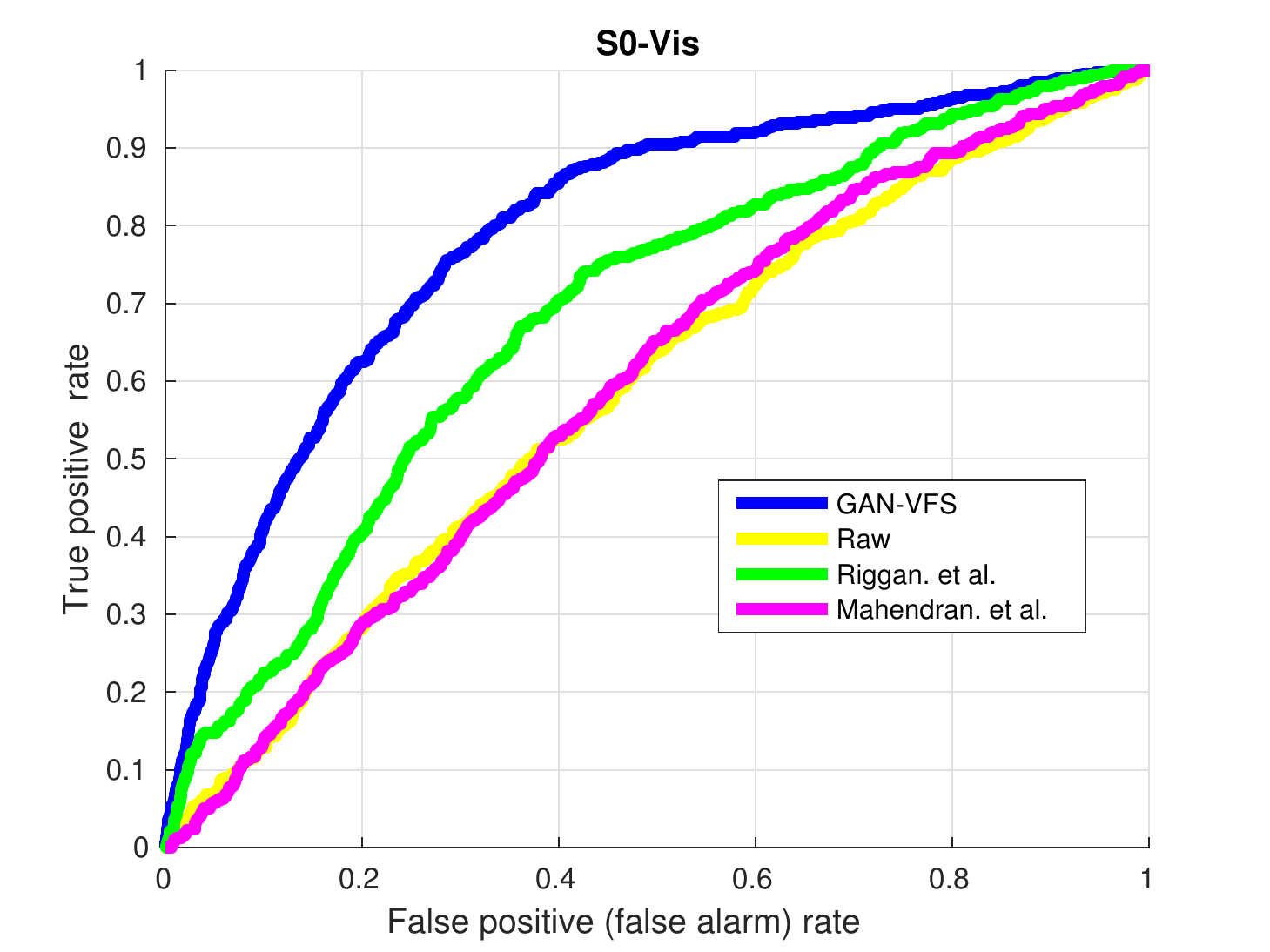}
 			\captionsetup{labelformat=empty}
 			\captionsetup{justification=centering}
 			\caption*{(d)}
 	\end{minipage}
 	\begin{minipage}{.4\textwidth}
 		\centering
 			\includegraphics[width=1\textwidth]{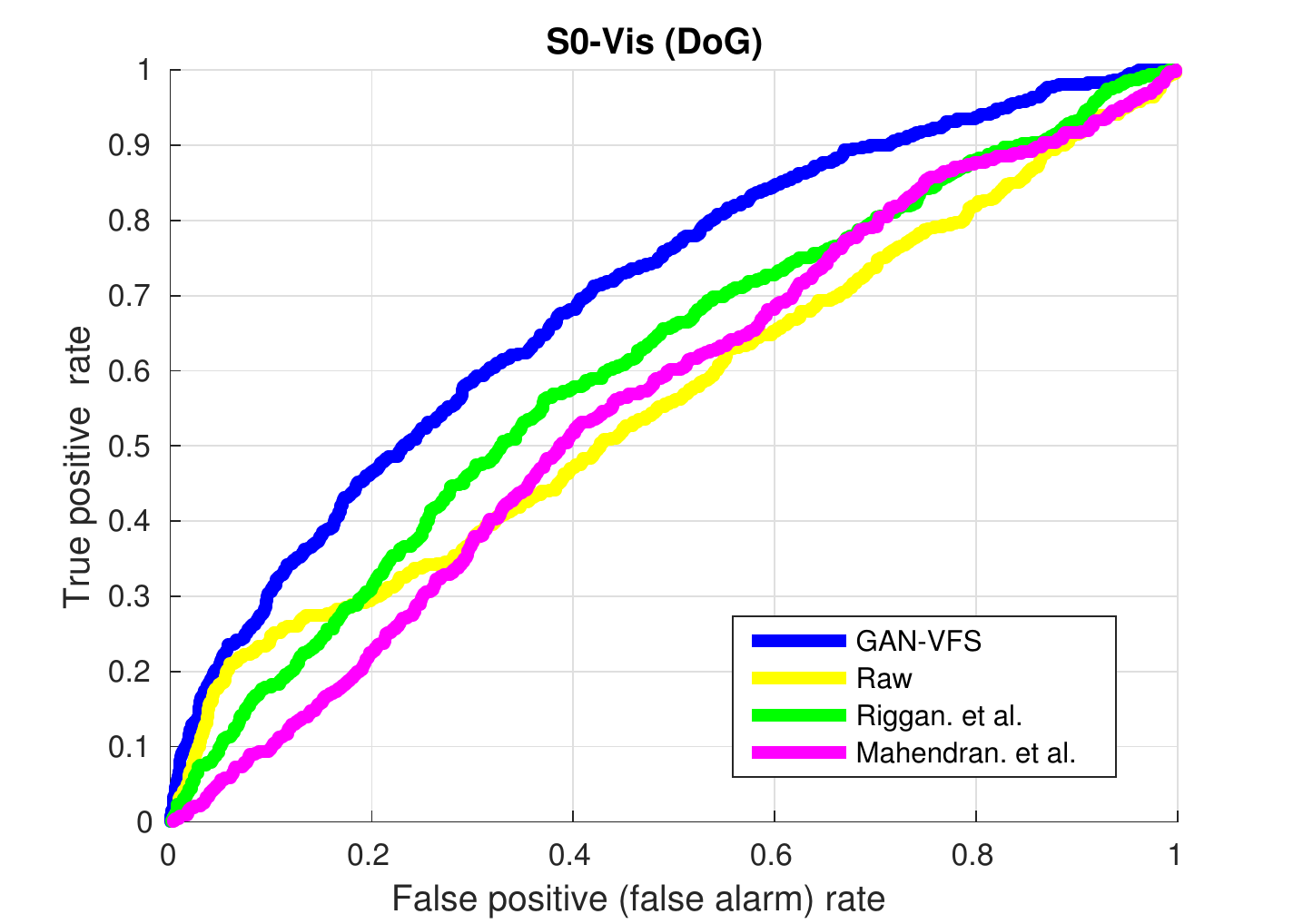}
 		\captionsetup{labelformat=empty}
 		\captionsetup{justification=centering}
 		\caption*{(e)}
 	\end{minipage}
	 	\vskip -8pt\caption{The verification results corresponding to different methods in terms of the ROC curves. (a) Polarimetric to Visible.  (b)  Polarimetric to Visible (DoG). (c)  S0 to Visible. (d)  S0 to Visible (DoG).} \label{fig:roc}
 \end{figure*}

\subsubsection{Face Verification Results}
As the goal of synthesizing the visible image given the  polarimetric image is to improve the verification results both for human examiners and computer vision algorithms, it is also important  to evaluate how the recovered visible images work in face verification tasks.   In this section,  we propose to use the performance of face verification as a metric to evaluate the polarimetric-to-visible face synthesis algorithms. The experimental details are as follows:  The ground truth images from all the 30 testing subjects are regarded as the  gallery set and the reconstructed visible images from the corresponding polarimetric images in the same 30 subjects are regarded as the probe set.  All the verification results are evaluated on the deep features extracted from the VGG-face model \cite{vggface} without fine-tuning.  Figure~\ref{fig:roc} plots the ROC curves corresponding to four experimental protocols.   The area under the curve (AUC)  and equal error rate (EER) results are reported  in Table~\ref{err}. From these results, it can be observed that the proposed method achieves the best performance evaluated on all four protocols. Meanwhile, it can be also observed that the reconstruction of the visible images from S0-Vis and Polar-Vis achieve  better verification results compared to their corresponding DoG versions.   This is mainly due to the fact that while extracting VGG features from the DoG filtered images, the network was not fine-tuned based on DoG images.   Therefore, features extracted for S0-Vis and Polar-Vis are better than features for their corresponding DoG versions.

\begin{table}[htp!]
\centering
\caption{The average EER and AUC results corresponding to different methods.}
\label{err}
\resizebox{0.48\textwidth}{!}{%
\begin{tabular}{|c|c|c|c|c|c|}
\hline
 & \textbf{} & Raw & \begin{tabular}[c]{@{}c@{}}Mahendran\\ \emph{et al.} \cite{inverting_cnn}\end{tabular} & \begin{tabular}[c]{@{}c@{}}Riggan\\\emph{et al.} \cite{btas_2016}\end{tabular} & \textbf{GAN-VFS} \\ \hline
\multirow{4}{*}{{AUC}} & S0-Vis & 58.64\% & 59.25\% & 68.52\% & \textbf{79.30\%} \\ \cline{2-6} 
 & Polar-Vis & 50.35\% & 58.38\% & 75.83\% & \textbf{79.90\%} \\ \cline{2-6} 
 & S0-Vis (DoG) & 56.09\% & 57.04\% & 60.57\% & \textbf{70.15\%} \\ \cline{2-6} 
 & Polar-Vis (DoG) & 59.54\% & 59.02\% & 67.31 & \textbf{70.41\%} \\ \hline
\multirow{4}{*}{{EER}} & S0-Vis & 43.96\% & 43.56\% & 34.36\% & \textbf{27.34\%} \\ \cline{2-6} 
 & Polar-Vis & 48.96\% & 44.56\% & 33.20\% & \textbf{25.17\%} \\ \cline{2-6} 
 & S0-Vis (DoG) & 46.48\% & 47.21\% & 41.27\% & \textbf{36.25\%} \\ \cline{2-6} 
 & Polar-Vis (DoG) & 43.33\% & 43.04\% & 37.23\% & \textbf{34.58\%} \\ \hline
\end{tabular}%
}
\end{table}

\section{Conclusion}
	We presented a new GAN-VFS network for synthesizing photo realistic visible face images from the  corresponding polarimetric images. In contrast to the previous methods that regarded visible feature extraction and visible image reconstruction as two separate processes, we took a different approach in which these two steps are jointly optimized. 	Quantitative and qualitative experiments  evaluated on a real polarimetric-visible dataset demonstrated that the proposed method is able to achieve significantly better results as compared to the recent state-of-the-art methods. In addition, an ablation study was performed to demonstrate the improvements obtained by different losses in the proposed method.

\section*{Acknowledgement}
 This work was supported by an ARO grant W911NF-16-1-0126.

{\small
\bibliographystyle{ieee}
\bibliography{egbib}
}

\end{document}